\documentclass{article}
\usepackage{spconf,amsmath,graphicx,hyperref}

\title{ClonEval: An Open Voice Cloning Benchmark}

\name{Iwona Christop \qquad Tomasz Kuczyński \qquad Marek Kubis}
\address{
    Adam Mickiewicz University, ul. Uniwersytetu Poznańskiego 4, 61-614 Poznań, Poland
}

\begin{document}
%
\maketitle
\begin{abstract}
We present a new benchmark for voice cloning text-to-speech models. The benchmark consists of an evaluation protocol, an open-source library for assessing the performance of voice cloning models, and an accompanying leaderboard. The paper discusses design considerations and presents a detailed description of the evaluation procedure. The usage of the software library is explained, along with the organization of the leaderboard. The evaluation results of selected open-source models are reported.
\end{abstract}
\section{Introduction}
\label{sec:intro}

With the advent of speech synthesis models capable of cloning arbitrary voices such as VALL-E by Wang et al. \cite{wang2023neural} and Voicebox by Le et al. \cite{le2023voicebox}, and the growing number of open-source and open-weight variants and alternatives to these models, there is an increasing demand for developing reliable and reproducible methods to evaluate voice cloning capabilities. The reliable procedure for evaluating voice-cloning should not assume any particular architecture of the model under study. It should also enable assessment of models that are not publicly available without excessive effort. To make the procedure reproducible it should have a form of a standardized benchmark with reasonable defaults set for data being utilized and evaluation steps being conducted. In order to facilitate the comparison of models with each other the benchmark should adopt the measurements techniques that are already accepted in the field while filling any under-specified parameters that could impact the evaluation result in an unpredictable way.

To address the aforementioned needs, we propose \emph{ClonEval}, a reproducible benchmark for voice cloning that consists of:
\begin{enumerate}
    \item A deterministic evaluation protocol that sets defaults for data, metrics, and models to be used in the voice cloning assessment process.
    \item An open-source software library that can be used to evaluate voice cloning models in a reproducible manner.
    \item A public leaderboard that enables comparison of the models against each other.
\end{enumerate}

The benchmark utilizes speaker embeddings to measure similarity between reference and generated samples. Besides reporting overall similarity which is used to establish the ranking we also analyze how well particular emotions are transferred by models under study to get more fine-grained view on their performance in capturing nuances. For the purpose of illustrating the use of our benchmark, we determine and analyze scores for selected open-weight models that were available online at the time of writing.

\section{Related Work}
\label{sec:related-work}

Current methods of evaluating voice cloning models have significant limitations. Both Wang et al. \cite{wang2023neural} and Le et al. \cite{le2023voicebox} employed word error rate (WER) to evaluate the intelligibility of synthesized speech. However, this metric only captures text accuracy and does not offer insight into voice cloning quality. Additionally, Le et al. \cite{le2023voicebox} conducted a human evaluation and reported mean opinion score (MOS) for speaker quality and similarity. While informative, subjective  assessments like these can be inconsistent and difficult to reproduce.

In order to evaluate speaker similarity directly, both Wang et al. \cite{wang2023neural} and Le et al. \cite{le2023voicebox} used the WavLM-TDNN model to obtain speaker embeddings for the prompt and the generated audio, and calculate the similarity between them. Le et al. \cite{le2023voicebox} also introduced the Fr\'echet Speech Distance (FSD), an adaptation of Fr\'echet Inception Distance (FID), which calculates distribution-level similarity between real and synthesized speech using features extracted with wav2vec 2.0.

Seamless Communication et al. \cite{communication2023seamless} proposes a new approach through AutoPCP which is a neural model designed to predict the human-rated prosody similarity (PCP score) between two audio signals. However, relying solely on neural networks for evaluation can introduce bias or misalignment with human perception.

\begin{figure*}[htbp]
    \centering
    \includegraphics[width=0.9\textwidth]{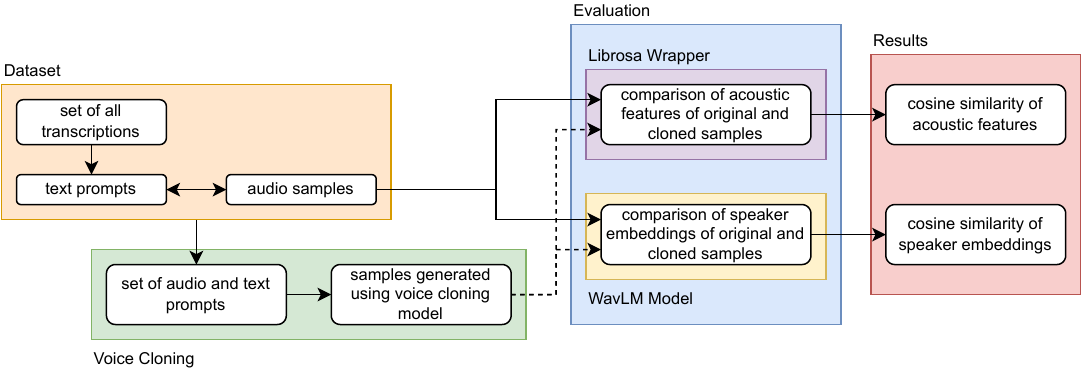}
    \caption{Overview of the evaluation process.}
    \label{fig:evaluation-process}
\end{figure*}

\section{Benchmark Design}
\label{sec:benchmark-design}

In order to develop a benchmark for voice cloning models, the following design principles were adopted:
\begin{enumerate}
    \item The voice cloning models should be treated as opaque, rather than assuming any particular architecture of the proposed solutions.
    \item There should be no need for human intervention in the evaluation process.
    \item The evaluation procedure should be easily adaptable to new models.
    \item The evaluation results should be reproducible.
\end{enumerate}

To fulfill principle (1), the requirements imposed on the evaluated model are reduced to an absolute minimum. The model must accept an audio sample of the voice to be cloned and a text prompt of an utterance as an input. The model is expected to generate a single audio sample as the output, representing the result of voice cloning. No additional assumptions, such as the ability to provide confidence scores or $n$-best lists, are made. While determining the mean opinion score (MOS) is a widely accepted practice for the subjective evaluation of TTS quality, adopting this practice in the context of an open benchmark would present significant challenges. Establishing an arena for comparing different models, as presented by Chiang et al. \cite{chiang2024chatbotarenaopenplatform}, and requesting people to vote would require curating the list of available models for evaluation. Such approach presents challenges, as it prevents independent researchers from evaluating models that are not publicly available due to commercial or ethical considerations. Relying on the results reported by the model authors could compromise the credibility of the findings, as verifying the accuracy of the reported results would require re-running the human evaluation, which is not feasible. Therefore, we have decided to adopt principle (2). Following the previous approaches \cite{wang2023neural,le2023voicebox}, we assess the similarity between the original and cloned samples using a pretrained WavLM \cite{chen2022wavlm} model, which utilizes its learned universal speech representations to achieve effective performance in numerous speech processing tasks in English. To fulfill principles (3) and (4) we developed \emph{ClonEval}, a software library that encapsulates the entire evaluation process. It can be used by vendors of voice cloning models to test their models independently and report their scores. The reported results can be reproduced with the use of the \emph{ClonEval} library by an external entity that has no access to the model, if the vendor releases cloned samples.

\begin{table*}[htbp]
  \centering
  \begin{tabular}{lcccccc}
    \hline
    \textbf{Dataset} & \textbf{OuteTTS} & \textbf{SpeechT5} & \textbf{VALL-E X} & \textbf{WhisperSpeech} & \textbf{XTTS-v2} \\
    \hline
    LS test-clean & 0.8836 & \textbf{0.9099} & 0.9010 & 0.9014 & 0.8881 \\
    \hline
    CREMA-D & 0.7359 & 0.7618 & 0.7412 & 0.7284 & \textbf{0.8060} \\
    RAVDESS & 0.7696 & 0.8265 & 0.7382 & 0.6972 & \textbf{0.8539} \\
    SAVEE & 0.5394 & 0.7987 & 0.7674 & 0.7725 & \textbf{0.8135} \\
    TESS & 0.8207 & \textbf{0.8521} & 0.7832 & 0.8188 & 0.8167 \\
    \hline
    \textbf{Average} & 0.7499 & 0.8298 & 0.7862 & 0.7837 & \textbf{0.8356} \\
    \hline
\end{tabular}
\caption{Cosine similarity between speaker embeddings of the original and cloned samples generated by the WavLM model.}
\label{tab:results-overall}
\end{table*}

\section{Evaluation Procedure}
\label{sec:evaluation-procedure}

The evaluation procedure outlined in Figure~\ref{fig:evaluation-process} involves two stages. First, samples are generated using a voice cloning model, with the primary dataset for cloning being LibriSpeech test-clean \cite{librispeech2015}. This dataset is currently one of the most widely used for model evaluation in many English speech processing tasks. Four additional popular English emotional speech datasets were selected to investigate the impact of emotionality on voice cloning quality. The selected emotional datasets were CREMA-D \cite{crema-d}, RAVDESS \cite{ravdess}, SAVEE \cite{savee}, and TESS \cite{tess}. As previously mentioned, the model must take as input a sample of the voice to be cloned and a text sample of an utterance. To obtain the text samples, for each voice sample, one text prompt was drawn from all those available in the given dataset, excluding the prompt corresponding to that sample.

Following the generation of samples through the voice cloning model, an evaluation is conducted using the WavLM model, with each sample undergoing identical processing. Given that the WavLM model accepts audio samples with a sampling rate of 16 kHz as input, the audio is resampled to this value. Speaker embeddings are then generated using the WavLM model.\footnote{\texttt{microsoft/wavlm-base-plus-sv}} For each pair of samples (reference and generated), the cosine similarity between their speaker embeddings from WavLM is calculated. The similarity values obtained for all samples from a given dataset are averaged to obtain the final evaluation result.

For the purpose of conducting fine-grained error analysis, we also extract acoustic features from each sample with Librosa \cite{mcfee2015librosa}.

\section{Software Library}
\label{sec:software-library}

The evaluation code and setup guidelines are provided in the GitHub repository\footnote{\url{https://github.com/amu-cai/cloneval}}. The procedure operates on two directories of input files, one with reference samples and the other with generated samples, matched by a filename. If emotional states are to be considered, filenames should also include the name of the corresponding emotion, allowing the results to be aggregated accordingly. The evaluation script produces two output files -- one reporting detailed metrics for each file pair and another summarizing averaged results, organized by emotion if relevant.

\section{Leaderboard}
\label{sec:leaderboard}

The evaluation results obtained with the \emph{ClonEval} library are presented on the Open Voice Cloning Leaderboard\footnote{\url{https://huggingface.co/spaces/amu-cai/Open\_Voice\_Cloning\_Leaderboard}}. The leaderboard shows the overall performance of models based on WavLM, as well as the scores for each emotional state and the average cosine similarity of the selected acoustic features. In addition, the community can submit voice cloning models via the designated tab.

\section{Experiments}
\label{sec:experiments}

\subsection{Overall}
\label{ssec:overall}

The results, presented in Table~\ref{tab:results-overall}, demonstrate the average cosine similarity between speaker embeddings from WavLM extracted from the reference sample and generated by each model. The XTTS-v2 \cite{xttsv2} achieved the highest score across most of the datasets, also resulting in the overall highest score. On the LS test-clean and TESS datasets, the SpeechT5 model \cite{ao2022speecht5} demonstrated superior performance. The OuteTTS-0.2-500M \cite{outetts}, VALL-E X\footnote{An open-source implementation was used. Available at \url{https://github.com/Plachtaa/VALL-E-X}} \cite{zhang2023vallex}, and WhisperSpeech \cite{whisperspeech} models exhibited slightly lower scores. It is noteworthy that in most cases the cosine similarity value exceeded 0.7, indicating that all of the models effectively cloned the voice from the reference sample to generate similar speaker embeddings. The best results were obtained for the LS test-clean dataset, suggesting that the models perform better in cloning non-emotional speech.

\subsection{Acoustic Features}
\label{ssec:acoustic-features}

Table~\ref{tab:results-features} presents the average cosine similarity between the acoustic features extracted from the reference sample and those generated by each model. The OuteTTS model demonstrated the highest performance, though the results obtained by the other models were not significantly different in most cases.

\begin{table*}[htbp]
  \centering
  \begin{tabular}{p{0.3\textwidth}ccccc}
    \hline
    \textbf{Feature} & \textbf{OuteTTS} & \textbf{SpeechT5} & \textbf{VALL-E X} & \textbf{WhisperSpeech} & \textbf{XTTS-v2} \\
    \hline
    pitch & \textbf{0.6094} & 0.5278 & 0.5818 & 0.5863 & 0.5287 \\ 
    mel spectrogram & \textbf{0.9259} & 0.9109 & 0.9208 & 0.9091 & 0.8622 \\ 
    RMS & \textbf{0.6970} & 0.6040 & 0.6810 & 0.6400 & 0.6238 \\ 
    spectral centroid & 0.7674 & \textbf{0.7855} & 0.7608 & 0.7485 & 0.7387 \\ 
    spectral flatness & \textbf{0.3229} & 0.3091 & 0.3199 & 0.2655 & 0.2418 \\ 
    spectral roll-off & 0.8198 & \textbf{0.8357} & 0.8134 & 0.8037 & 0.7907 \\ 
    tempogram & \textbf{0.5159} & 0.5127 & 0.5030 & 0.5071 & 0.3658 \\ 
    chromagram & 0.6036 & \textbf{0.6312} & 0.5966 & 0.5694 & 0.5775 \\ 
    pseudo-constant-Q transform & \textbf{0.6707} & 0.6261 & 0.6499 & 0.6388 & 0.6242 \\ 
    constant-Q chromagram & 0.7167 & \textbf{0.7330} & 0.7146 & 0.6836 & 0.7138 \\ \hline
    WavLM & 0.7499 & 0.8298 & 0.7862 & 0.7837 & \textbf{0.8356} \\
    \hline
\end{tabular}
\caption{Cosine similarity between the values of selected acoustic features of the original and cloned samples obtained for LS test-clean dataset.}
\label{tab:results-features}
\end{table*}

\begin{table*}[htbp]
  \centering
  \begin{tabular}{lccccc}
    \hline
    \textbf{Emotion} & \textbf{OuteTTS} & \textbf{SpeechT5} & \textbf{VALL-E X} & \textbf{WhisperSpeech} & \textbf{XTTS-v2} \\
    \hline
    anger & 0.7197 & 0.7923 & 0.7623 & 0.7462 & \textbf{0.8098} \\
    disgust & 0.7034 & 0.8172 & 0.7600 & 0.7458 & \textbf{0.8325} \\
    fear & 0.6953 & \textbf{0.7996} & 0.7466 & 0.7601 & 0.7929 \\
    happiness & 0.7329 & 0.8068 & 0.7658 & 0.7462 & \textbf{0.8160} \\
    neutral & 0.7370 & 0.8322 & 0.7699 & 0.7714 & \textbf{0.8480} \\
    sadness & 0.7135 & 0.8099 & 0.7525 & 0.7516 & \textbf{0.8365} \\
    \hline
    \textbf{Average} & 0.7170 & 0.8097 & 0.7595 & 0.7536 & \textbf{0.8226} \\
    \hline
  \end{tabular}
  \caption{Cosine similarity between speaker embeddings of the original and cloned samples generated by the WavLM model for emotional states.}
  \label{tab:results-emotions}
\end{table*}

The similarity of the pitch suggests that both samples share overall trends but exhibit noticeable differences in details, which indicates that they could be spoken by the same speaker, but the textual content potentially differs, a conclusion that aligns with the principles of the evaluation procedure. Additionally, the presence of distortions in the generated sample could potentially reduce the value of similarity.

The high similarity of the mel-spectrogram indicates that the recordings are highly aligned in a perceptual scale, suggesting that the recordings were produced by the same speaker. The high level of similarity observed for both the chromagram and the constant-Q chromagram indicates that the samples share moderately similar harmonic structures, suggesting variations in spoken content with shared harmonic trends.

The values of similarity for spectral features, including centroid, flatness, and roll-off, further indicate that the speakers are highly similar in frequency distribution and that the noise is minimal.

The value of similarity for RMS indicates a uniform loudness, along with variations in timing, emphasis, and intensity, suggesting that the signals may be from the same speaker, though with different content. The observed similarity of the tempogram indicates that there is partial alignment in the rhythmic structures, suggesting that the samples likely represent different content.

The similarity value for pseudo-constant-Q transform further suggests similar speaking styles with some variations in content.

As demonstrated above, the values of similarity for all considered acoustic features indicate that the samples were spoken by the same individual, but the textual content was different. This finding is consistent with the similarity obtained from WavLM.

\subsection{Emotions}
\label{ssec:emotions}

Table~\ref{tab:results-emotions} presents the average cosine similarity scores for each emotional state. XTTS-v2 achieved the best results for nearly all emotions, with only a slight difference for \emph{fear}. OuteTTS, on the other hand, consistently performed the worst. All models were most effective at cloning the neutral state and least effective at cloning highly expressive emotions, such as \emph{fear}, \emph{anger}, and \emph{disgust}.

\section{Conclusion}
\label{sec:conclusion}

We introduced \emph{ClonEval}, a standardized and reproducible benchmark for evaluating voice cloning systems to address the growing need for automatic, reliable and architecture-agnostic evaluation methods. By combining a deterministic evaluation protocol, an open-source software library, and a public leaderboard, \emph{ClonEval} provides a comprehensive framework for assessing voice cloning models. Our benchmark builds upon widely recognized evaluation techniques while introducing well-defined defaults to minimize ambiguity and enhance reproducibility. 

The preliminary results reported in Section~\ref{sec:experiments} show that the performance of the models under study vary when we consider individual emotions. While all the models performed best when cloning emotionally neutral utterances, their behavior differed with respect to \emph{fear}, \emph{anger} and \emph{disgust}. The majority of the evaluated models under-performed while cloning utterances that included \emph{fear}.

As part of our future work, we plan to expand our evaluation framework by applying it to a wider variety of voice cloning models and diverse datasets. Additionally, we plan to incorporate human evaluation to complement objective metrics with subjective judgments. This will provide insights into the perceived quality and naturalness of the generated audio samples.

\bibliographystyle{IEEEbib}
\bibliography{bibliography}

\end{document}